\documentclass[letterpaper, 10 pt, conference]{ieeeconf}  

\IEEEoverridecommandlockouts                              

\usepackage{graphics} 
\usepackage{graphicx}
\usepackage{amsmath} 
\usepackage{amssymb}  
\usepackage{booktabs}
\usepackage{subcaption}
\usepackage{xcolor}
\usepackage{hyperref}

\title{\LARGE \bf
Leveraging Multimodal Haptic Sensory Data for Robust Cutting
}

\author{Kevin Zhang$^{1}$, Mohit Sharma$^{1}$, Manuela Veloso$^{2}$, and Oliver Kroemer$^{1}$
\thanks{*This work was supported by the Sony Corporation. This work was in part supported by the Office of Naval Research under Grant No. N00014-18-1-2775.}
\thanks{$^{1}$Kevin Zhang, Mohit Sharma, and Oliver Kroemer are with The Robotics Institute,
        Carnegie Mellon University,
        Pittsburgh, PA 15213, USA
        {\tt\small \{klz1, mohits1, okroemer\}@cs.cmu.edu}}%
\thanks{$^{2}$Manuela Veloso is with the Machine Learning Department, Carnegie Mellon University,
        Pittsburgh, PA 15213, USA
        {\tt\small mmv@cs.cmu.edu}}%
}

\begin{document}

\maketitle
\thispagestyle{empty}
\pagestyle{empty}

\begin{abstract}
Cutting is a common form of manipulation when working with divisible objects such as food, rope, or clay. Cooking in particular relies heavily on cutting to divide food items into desired shapes. However, cutting food is a challenging task due to the wide range of material properties exhibited by food items. Due to this variability, the same cutting motions cannot be used for all food items. Sensations from contact events, e.g., when placing the knife on the food item, will also vary depending on the material properties, and the robot will need to adapt accordingly. 

In this paper, we propose using vibrations and force-torque feedback from the interactions to adapt the slicing motions and monitor for contact events. The robot learns neural networks for performing each of these tasks and generalizing across different material properties. By adapting and monitoring the skill executions, the robot is able to reliably cut through more than 20 different types of food items and even detect whether certain food items are fresh or old.


\end{abstract}

\section{INTRODUCTION}

Cutting and preparing ingredients is a fundamental part of cooking, but it is also a monotonous and time-consuming task for chefs. Having robots help prepare meals by cutting food would thus save time and also encourage healthier eating habits.
However, cutting food is a challenging problem because food items differ greatly in both visual properties (e.g,  shape, color, and texture) and mechanical properties (e.g., hardness, density, and friction) \cite{sahin2006physical}. 
To cut a wide range of food items, the robot will need to generalize its cutting skills across these mechanical properties. In particular, the robot needs to reliably detect key contact events (e.g., hitting the food item or the cutting board) despite material variations, and it needs to adapt its slicing motion to the individual food items (e.g., applying more downward force and less lateral motion to cut a cucumber instead of a tomato).

\begin{figure}[t]
    \centering
    \includegraphics[width=0.47\textwidth]{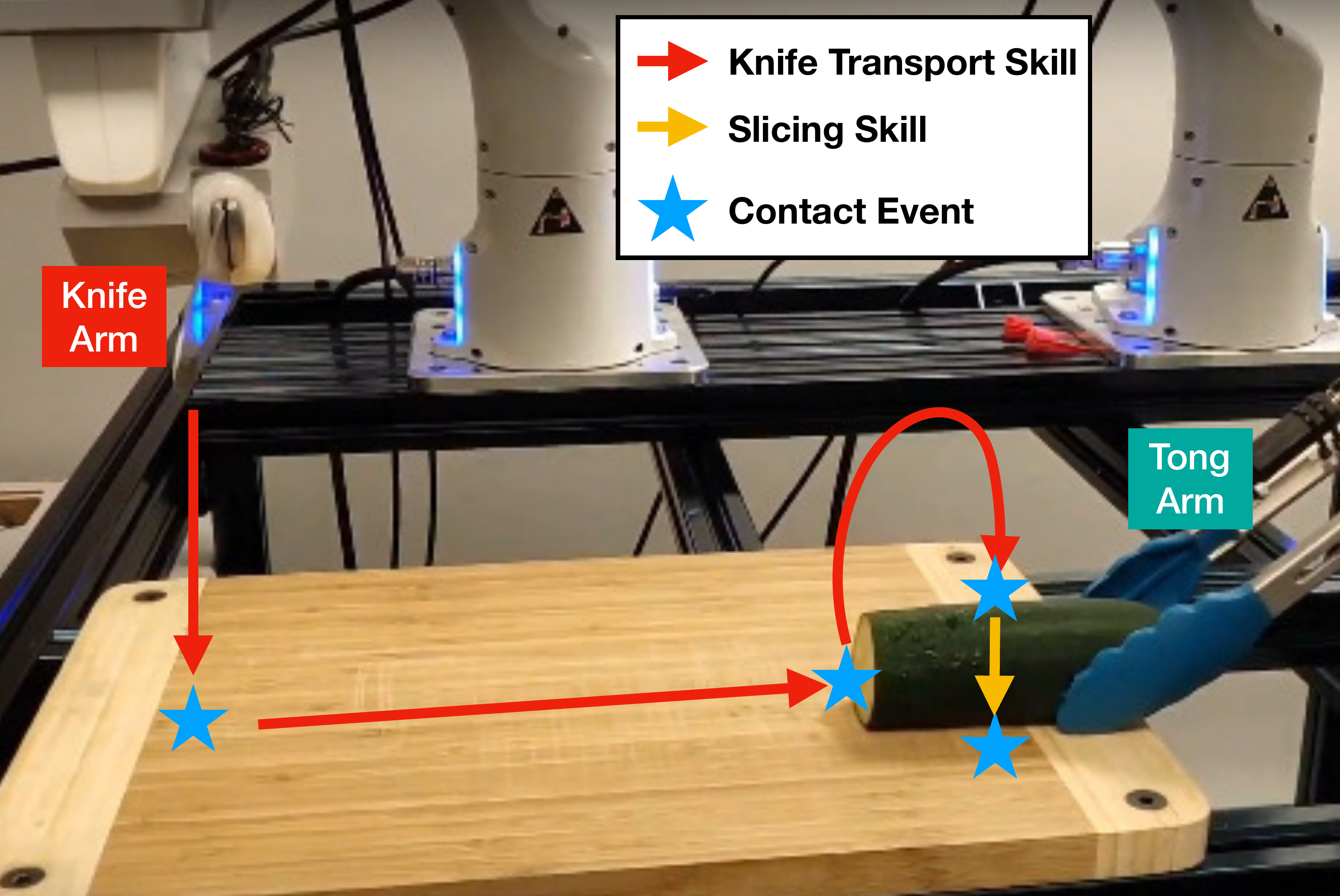}
    \caption{Illustration of the skill sequence for cutting a slice. The robot needs to detect the contact event and adapt the slicing skill to the material properties of the food item. Failure events, e.g., knife slipping from the food item, are not shown but must also be detected. The final red and yellow skills can be repeated to cut additional slices.}
    \label{fig:knife_diagram}
    \vspace{-1.25em}
\end{figure}

We divide the cutting task into multiple low-level skills that are then sequenced together based on the detected contact events. 
Fig. \ref{fig:knife_diagram} shows an example sequence of skills for positioning the knife and cutting a slice off of a cucumber. To perform this task in a robust manner, the robot learns a set of neural networks to detect the contact events and adapt the slicing skill to the food's properties.

The robot uses tactile and haptic feedback to detect the contact events and the material properties. The robot uses its joint torque sensors to estimate the forces and torques being applied to the knife. For dynamic tactile sensing, the knife, tongs, and cutting board are equipped with contact microphones. These microphones can detect the vibrations caused by contact events and by the knife cutting through the food. This approach was inspired by the fast afferents located in human skin, which detect vibrations during manipulation tasks and tool usage \cite{johansson2009coding}. Although the robot is equipped with RGB-D cameras, we do not use them in this paper. Interactive perception and tactile sensing are often more reliable for estimating material properties (e.g., differentiating between a fresh and an old cucumber) as well as detecting contact events (e.g., differentiating between contact and slightly before contact).

Given the tactile and haptic data, the robot learns a set of neural networks for identifying the contact events and the material properties. This data-driven approach allows the robot to automatically extract high-level features for generalizing across the different types of food items. The learned networks are subsequently integrated into the robot's finite state machine for sequencing skills. The material type is predicted from initial contact. The type is then mapped to parameters for adapting the slicing skill, which is modeled as a parameterized dynamic motor primitive with impedance control. 
The proposed approach was successfully implemented on a bimanual Franka robot setup.

\section{RELATED WORK}

Previous work on robotic cutting has focused on model-predictive control (MPC) to generalize between different types of food items. Lenz et al. \cite{lenz2015deepmpc} proposed learning a deep network to model the cutting interaction's dynamics for their DeepMPC approach. This network continually estimated the latent material properties throughout the cutting process using interactive perception and a recurrent network structure. Mitsioni et al. \cite{deepmpcionna2019} extended this approach by incorporating measurements from a force-torque sensor into the dynamics model. They again used a recurrent neural network to model the dynamics of the cutting task. By contrast, rather than attempting to model the complex dynamics of cutting interactions in detail, we instead directly learn a dynamic motor primitive (DMP) policy for the cutting skill. The robot learns the DMP from human demonstrations, which allows the human to implicitly transfer prior knowledge to the robot in an intuitive manner (e.g., using long smooth lateral motions for cutting). The DMP adapts to different food items based on two input parameters, which we estimate based on the initial interactions with the food item. Learning the policy is generally easier than learning the low-level dynamics for the cutting task. Unlike the MPC approaches, we do not continuously update our parameters to material variations during individual cuts.

A large part of robotics research has focused on using vision as the primary sensory modality for performing manipulation tasks \cite{levine2016end}, \cite{levine2018learning}, \cite{pinto2016supersizing}. However, for complex and contact rich tasks, such as food cutting, visual sensory data alone is insufficient since there can be major occlusions and visually ambiguous scenarios. For instance, the knife may have cut the food item completely, but the new slice may not have fallen down. 

Many previous works have used haptic feedback to accomplish a range of robotics tasks. For instance, \cite{drimus2011classification} use haptic feedback to classify both deformable and rigid objects, \cite{chu2015robotic} use haptic feedback to classify haptic adjectives. Furthermore, haptic feedback has also been used to infer the object properties of deformable objects \cite{frank2010learning, takamuku2007haptic, kaboli2014humanoids}. Gemici and Saxena \cite{gemici2014learning} use tactile feedback along with other robot data, such as poses, to determine the physical properties of different food items. 
They use specific tools and actions to infer carefully designed features which are then used to predict properties, such as the elasticity and hardness of the food items, by training a network using supervised learning.
In contrast, instead of determining the exact material properties of food items, we directly detect the food material which is then mapped to parameters that correspond to performing the slicing skill for that food item.

The use of vibration signals for robotic tasks has been explored in the past. Clarke et al. \cite{clarke2018learning} estimate the weight of materials that were scooped by a Sawyer arm using a contact microphone and a shaking motion. They test their algorithm on granular materials such as coffee beans and rice. 
Additionally, \cite{liang2019making} use a microphone to estimate the height of the liquid in a container that a robot poured into.
\cite{kroemer2011learning} use a contact microphone at the end of a robot's end-effector to learn a material's properties by stroking and visually inspecting it.
Previous works have also focused on using tactile feedback for object classification \cite{bhattacharjee2012haptic} for both rigid and deformable objects.
Similar to the above works, we use the rich auditory signals generated during robot cutting to robustly cut food items. However, in contrast to the above approaches, we combine the continuous auditory data to detect and infer the occurrence of both discrete and continuous events during deformable object manipulations. We use these inferred events to chain the different skills that are required to cut different food items.

\section{Overview}
Cutting is a challenging problem because there are numerous events that can occur as the robot is executing the cutting motion. Our goal is to learn the low-level skill primitives required to perform the cutting task as well as monitor for contact events using sensory feedback.

\subsection{Food Cutting Setup}

\vspace{-0.5em}

\begin{figure}[thpb]
    \centering
    \includegraphics[width=0.48\textwidth]{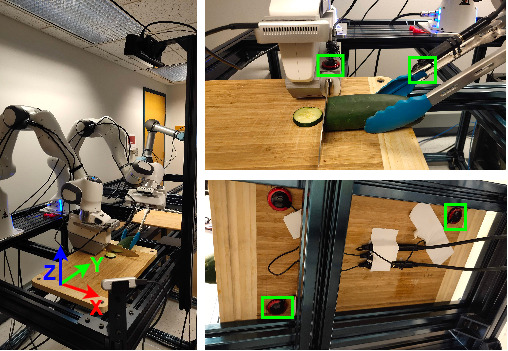}
    \caption{Experimental setup with contact microphones shown in green boxes.}
    \label{fig:experimental_setup}
    \vspace{-0.5em}
\end{figure}

Our robot setup for collecting cutting data is shown in Fig. \ref{fig:experimental_setup}. The setup is used to collect both the training data of different cutting events and evaluate the learned networks. Two Franka Panda Arms are mounted next to each other with overlapping workspaces. The left Franka Arm is grasping a knife with a 3D printed handle to provide a firm grip. We refer to this arm as the Knife Arm. The right Franka Arm has 9 inch tongs bolted onto its hand as fingers. We refer to this arm as the Tong Arm.

There are four contact microphones attached to different objects in the setup. One is located on the knife handle, and it is able to sense vibrations from the blade of the knife. Another is located on the right tong of the Tong Arm, and it is able to sense vibrations from both tongs. Finally, two microphones are located underneath the cutting board in opposite corners. The vibration signals from the microphones are captured using a Behringer UMC404HD audio interface system. The robot's position, forces, and vibration data are all synchronized and collected using ROS \cite{quigley2009ros}. 

\subsection{Cutting Process}

\begin{figure}[t]
    \centering
    \includegraphics[width=0.48\textwidth]{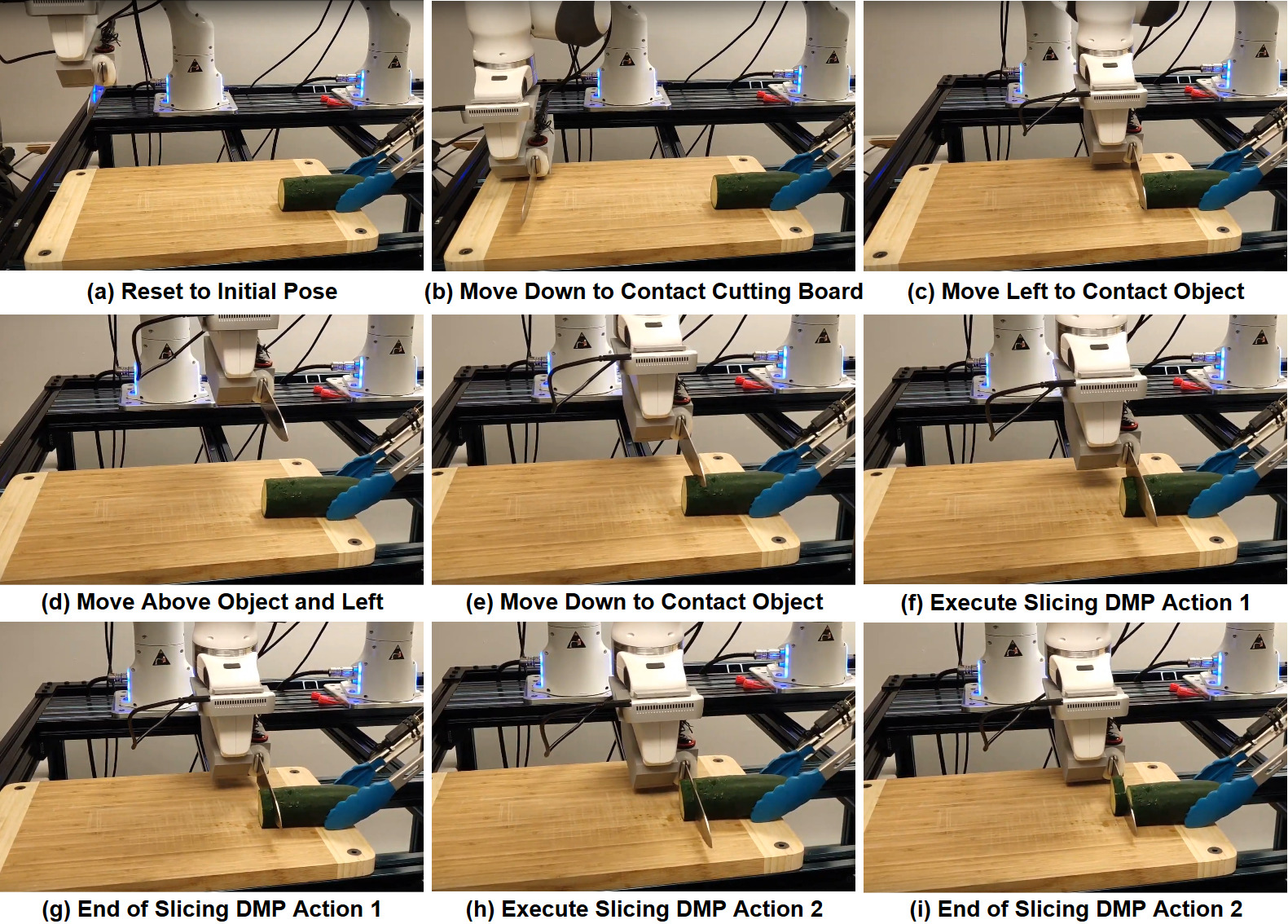}
    \caption{Time-lapse of Slicing a Cucumber}
    \label{fig:cutting_diagram}
    \vspace{-0.5em}
\end{figure}

Fig. \ref{fig:cutting_diagram} illustrates the key events of the cutting process. 
The Knife Arm starts at its initial pose \textbf{(a)}. It then moves down until contact to localize the cutting board \textbf{(b)}. To avoid scraping the cutting board, the robot lifts the knife slightly. The robot then moves the knife left until contact, thus implicitly localizing the end of the object without using vision \textbf{(c)}. Next, the robot lifts the knife up above the object and moves the blade left the desired thickness of the first cut \textbf{(d)}. The Knife Arm moves down until it contacts the object \textbf{(e)}. The sensory data from this downward motion into the food item is used to determine the appropriate cutting parameters for the object. With the cutting parameters inferred, the Knife Arm executes the slicing skill until the robot has cut through the object completely and made contact with the cutting board \textbf{(f)-(i)}. The robot loops through steps \textbf{(d)-(i)} until the desired number of slices has been cut.

\section{Multimodal Event Monitoring}\label{sec:multi_modal_event_monitoring}

\begin{figure}[t]
    \centering
    \includegraphics[width=0.48\textwidth]{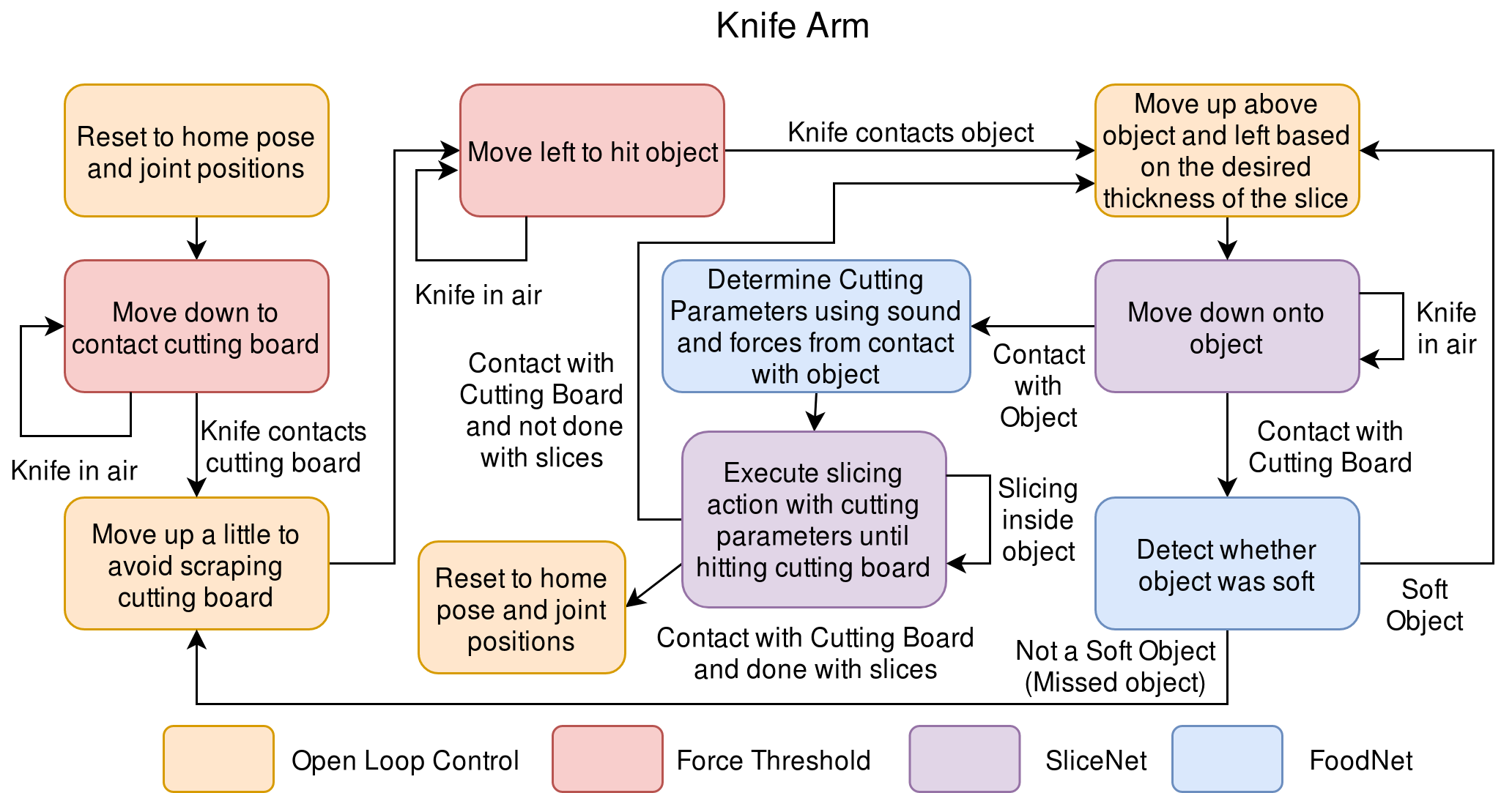}
    \caption{Knife Arm Slicing Flow Diagram}
    \label{fig:knife_arm_flow_diagram}
    \vspace{-0.5em}
\end{figure}

Fig.~\ref{fig:knife_arm_flow_diagram} shows the finite state machine for sequencing the different skill primitives used in the cutting process. To transition between the different states of the finite state machine, the robot needs to monitor the sensory feedback from the environment and determine when to terminate each skill. The termination conditions associated with the different skills are listed in Table~\ref{table:skill_info_table}. 

\begin{table}[t]
    \centering
    \begin{tabular}{@{}ll@{}}
Skill & Termination Conditions \\ \midrule
Move down on cutting board & Contact on $Z$-axis      \\
Move left to hit object & Contact on $X$-axis      \\
Move down onto object & Hitting Event Monitoring      \\
Slicing Action & Slicing Event Monitoring 
\end{tabular}
      \captionof{table}{Different skills used in the cutting process and their respective termination conditions.}
      \label{table:skill_info_table}
      \vspace{-0.5em}
\end{table}

\begin{figure}[t]
    \centering
    \includegraphics[width=0.48\textwidth]{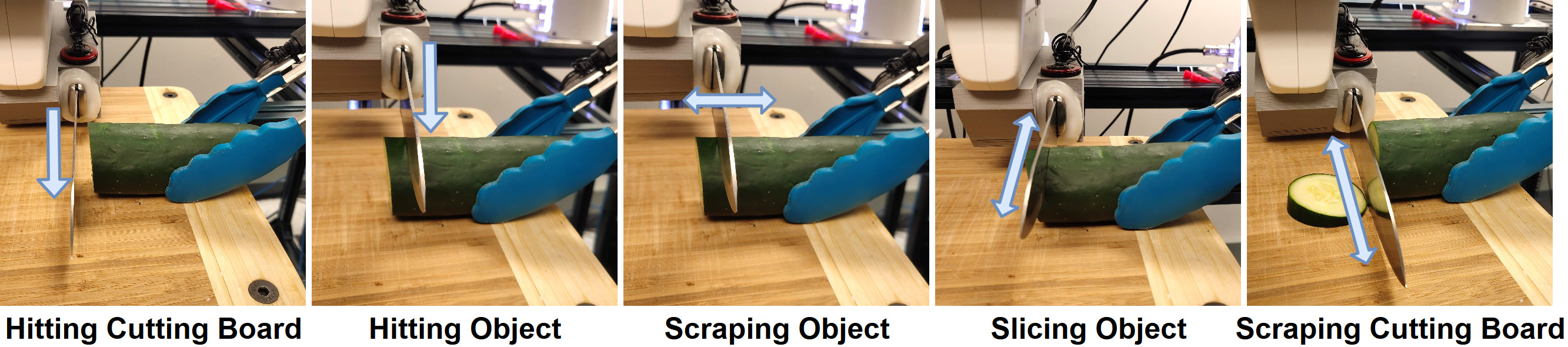}
    \caption{Different events that may occur when cutting.}
    \label{fig:robot_monitoring}
    \vspace{-1em}
\end{figure}

For the initial object and cutting board localization skills, simple directional force thresholds are sufficient for detecting the contact events. However, the moving-down-onto-the-object skill and the slicing skill have more advanced termination conditions due to the inherent variability when interacting with deformable objects of various material properties. Fig.~\ref{fig:robot_monitoring} illustrates the events that we monitor and detect during these skills.

The robot may hit the cutting board directly when trying to go down onto the object as either the object was too soft (e.g., tofu) or the knife slipped off of the object when the desired slice thickness was too thin.
The robot needs to detect these events quickly so that it does not unnecessarily execute slicing actions on the cutting board which tend to damage the cutting board, dull the knife, and waste time.

The knife may also slip on the surface of the object when it is executing the slicing action. This event sometimes occurs when the skin of the object is sloped and tough, like a watermelon's surface. If the robot does not notice the slipping event quickly, the resulting slice will have an undesired thickness. The robot can recover from this error by reorienting the knife to prevent slippage or by moving down harder on the object to firmly embed the knife within the object during the initial approach.



To continuously monitor and detect each of the events in Fig. \ref{fig:robot_monitoring}, we use the vibration feedback from the contact microphones and the force feedback sensed by the Knife Arm. We first process the multimodal data into features and labels (see section \ref{sec:multi-modal_data_processing}). We subsequently use the multimodal input features and event labels to train a fully connected Neural Network, called \textbf{SliceNet}. SliceNet consists of 
3 hidden layers with 100 units in each hidden layer, sigmoid activation at each layer, and 1 dropout layer before the last hidden layer. Training utilizes the categorical cross-entropy loss function with Adam optimizer for 50 epochs. SliceNet has 6 output classes: in air, hitting cutting board, hitting object, scraping object, slicing object, and scraping cutting board. The in air class is a background class where the knife is not in contact with any object. 

\section{Dynamic Slice Adaptation}

The slicing motion is the most important skill for the robot to learn for the cutting task. To enable the robot to learn the slicing skill, we use Dynamic Movement Primitives (DMPs), a general framework for smooth trajectory generation for complex actions \cite{schaal2006dynamic}. Our DMP formulation allows the robot to learn a back and forth motion which imitates a demonstrated slicing strategy. We chain multiple DMPs together to completely cut a slice from the object.

Utilizing the same slicing action for every food item is inefficient and error prone, as different food items have different physical and material properties. For instance, to cut soft tofu, the slicing skill should have a large vertical motion and a small horizontal motion. By contrast, a watermelon requires a much larger horizontal motion and a smaller downward motion. To adapt our slicing skill to different materials, we parameterize the DMPs using two additional input parameters that control the amplitude and height of the DMP. 
SliceNet  consists  of  3
hidden  layers  with  100  units  in  each  hidden  layer,  sigmoid
activation at each layer, and 1 dropout layer before the last
hidden  layer.  Training  utilizes  the  categorical
\subsection{DMP Formulation}

DMPs consist of linear dynamical systems which are learned for each skill component, i.e., the horizontal $X$ and vertical $Z$ axis movements of the knife in our case. We utilize a parameterized form of DMPs proposed by Kroemer and Sukhatme \cite{kroemer2016meta} which is defined as:
\begin{equation}\label{eq:dmp_1}
\ddot{y} = \alpha_z(\beta_z \tau^{-2}(y_0-y) - \tau^{-1}\dot{y}) + \tau^{-2}\sum_{j=1}^M \phi_j f(x;\boldsymbol{w}_j)
\end{equation}

where $y$ is the state, $y_0$ is the initial state, $\alpha_z$ and $\beta_z$ are constants that define the system's spring and damper coefficients, $\tau$ is a time coefficient, $x$ is the state of the canonical system, and $f$ is a forcing function. The canonical state $x$ acts as a timer for synchronizing multiple linear systems. It starts at $x=1$ and decays according to $\dot{x} = -\tau x$. $\phi_j$ are object features that allow the skill to adapt to different scenarios. $\boldsymbol{w}_j$ is a vector of weights $\in \mathbb{R}^K$ of the forcing function $f(x;\boldsymbol{w}_j)$, which shapes the DMP's trajectory. The forcing function $f$ is of the form:
\begin{equation}
f(x;\mathbf{w}_j) = \alpha_z\beta_z\left(\frac{\sum_{k=1}^K \psi_k(x)w_{jk} x}{\sum_{k=1}^K \psi_k(x)} + w_{i0}\psi_0(x) \right)
\end{equation}

where $w_{jk}$ is the $k$th element of the vector $\boldsymbol{w}_j$, $\psi_k(x) \ \forall k \in {1,...,K}$ are Gaussian basis functions, and $\psi_0$ is a basis function that follows a minimum jerk trajectory from 0 to 1. 

The above DMP formulation is amenable for the cutting process as it allows us to easily start the slicing motion from any arbitrary position, which further allows us to chain multiple DMPs together without any additional constraints. More importantly, since there is a separate DMP component for each axis of motion $(X$ and $Z)$, we can modify both the amplitude of the $X$-axis for forward and backward motions as well as the overall downward height displacement in the $Z$-axis. We do not modify the $Y$-axis since there is very little $Y$-axis motion when cutting.

We adapt the $X$ and $Z$ motions using the object features $\phi_j$ in the above formulation. We use two object features for every dimension, thus $j = 2$. We set $\phi_0$ to 1 as a source of bias for every object equally, while we parameterize $\phi_1$ based on the food material being cut. Since we have separate DMPs for both $X$ and $Z$-axes we set $\phi_1$ for each of the linear systems separately. 
We refer to each of the above parameters as the slicing action parameters, i.e. $\phi_1^x$ for the amplitude of the $X$-axis and $\phi_1^z$ for the height displacement. To adapt our slicing DMP to different food items, we learn to infer these parameters based on feedback while executing the moving-down-onto-the-object skill.

\subsection{Learning DMP Slicing Action Parameters}

\begin{figure}[thpb]
	\centering
	\includegraphics[width=0.48\textwidth]{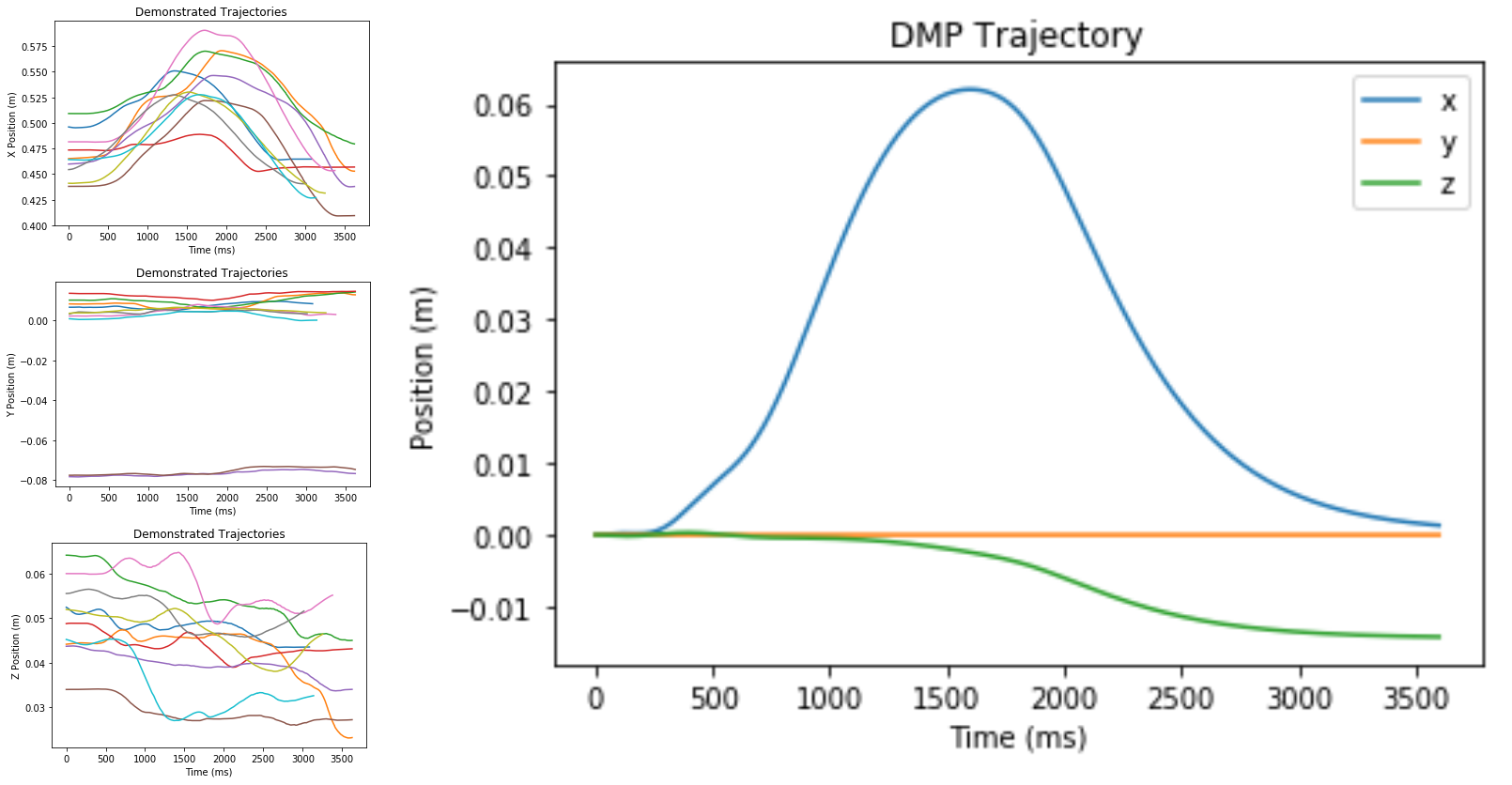}
	\caption{10 Demonstrated trajectories we collected on the left and learned DMP trajectory on the right.}
	\label{fig:dmp_trajectories}
\end{figure}

To learn the slicing DMP, we first need to learn the weights $\boldsymbol{w}_j$ of the forcing function as shown in \eqref{eq:dmp_1}. 
We use imitation learning (IL) to learn these DMP parameters from demonstrations. We use kinesthetic demonstrations to perform the robot slicing motion. Then, we use the saved trajectories from the demonstrations to learn the weights of the DMP trajectory for each axis using ridge regression.
Fig. \ref{fig:dmp_trajectories} shows the demonstrated trajectories and the smooth DMP trajectory generated by our learned DMP parameters for each of the three axes.

We additionally need to infer the appropriate slicing action parameters ($\phi_1^x$ and $\phi_1^z$) to adapt the slicing skill to the different food items. To achieve this goal, we collect the amplitude $\phi_1^x$ and height $\phi_1^z$ parameters for each class of objects manually, i.e., a human adapts the parameters for each training food item.
We then train a neural network which we refer to as \textbf{FoodNet} to infer the material class of an object based on sensory signals from the beginning of the slicing motion. Once we infer the material type, we substitute the parameters of the predicted material type into the slicing action parameters. This scheme allows us to adapt the same slicing motion to different food items to cut them more efficiently. 


\subsection{FoodNet Material Adaptation}

\label{sec:foodnet}

FoodNet's purpose is to classify the material of the food being cut and adapt the slicing skill accordingly. We processed the multi-modal data into features and labels, which we explain in detail in the next section. We then utilized a fully connected neural network with 3 hidden layers with 100 units in each hidden layer, sigmoid activation at each layer, and 1 dropout layer before the last hidden layer. Training utilized the categorical cross-entropy loss function with Adam optimizer. In total, there are 25 classes of objects. 

\section{Multi-Modal Data Processing}
\label{sec:multi-modal_data_processing}

\begin{figure}[thpb]
    \centering
    \includegraphics[width=0.48\textwidth]{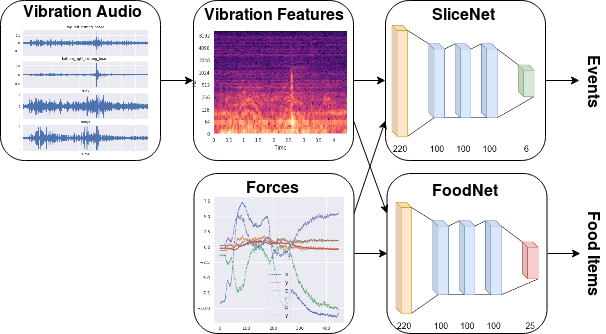}
    \caption{SliceNet and FoodNet System Diagram}
    \label{fig:overall_system_diagram}
    \vspace{-0.5em}
\end{figure}

The SliceNet and FoodNet architectures are depicted in Fig. \ref{fig:overall_system_diagram} above. Both of these networks use multimodal feedback from the environment 
i.e., the vibrational feedback from the microphones and the force feedback from the Knife Arm. We use early fusion and combine the multi-modal data together before sending them as input to the networks.




The first critical component of our multimodal data processing system is the vibration processing to retrieve the high-frequency feedback during cutting. We capture the sound from the 4 microphones by reading in the raw sound data using the python-sounddevice package \cite{sound_device}. 
We sample from the microphones with a 44.1kHz frequency, and we use an audio buffer to process 0.1 seconds of sound at a time.

We use Librosa \cite{mcfee2015librosa} for processing the vibration signals and to extract a wide range of audio features. We use Mel-frequency cepstral coefficients (MFCCs), chromagrams, mel-scaled spectrograms, spectral contrast features, and tonal centroid features (tonnetz). The Mel frequency scale provides a rough model of human frequency perception \cite{stevens1937scale}. Mel-frequency cepstral coefficients are often used for speech recognition systems and to represent timbre  \cite{logan2000mel}. Chromagrams project audio from the entire spectrum onto 12 bins representing the 12 distinct semitones of the musical octave \cite{ellis2007chroma}. 
Spectral Contrast features have been shown to perform well when discriminating between different music types \cite{jiang2002music}. Tonal centroid features detect changes in the harmonic content of audio signals using chroma features \cite{harte2006detecting}. 

We process each audio channel separately and extract the mean of each feature over the 0.1 second window. There are 40 Mel-frequency cepstral coefficients, 12 chromagram features, 128 mel-scaled spectrogram features, 7 spectral contrast features, and 6 tonal centroid features. Thus in total, we have 193 features per channel, and 772 features for the 4 channels every 0.1 seconds. However, given the ablation experiment in Section \ref{sec:ablation_study}, we utilize only Mel-frequency cepstral coefficients from each microphone for SliceNet and FoodNet in all the other experiments.

In addition to the vibrational features, we also make use of the force feedback provided by the robot. The robot provides us with force feedback at a 1kHz frequency; however, we subsample it to get a reduced 100 Hz frequency that is communicated over ROS. To match up the forces with the 0.1 seconds of sound, we use a buffer of the last 10 robot forces which are with respect to the x, y, z, roll, pitch, and yaw axes. In total there are 60 force features that are concatenated with the 772 vibration features resulting in a total of 832 features.

\section{Data Collection}

We collected a comprehensive dataset of a variety of different materials to train FoodNet. We also collected a dataset of events that may occur during cutting to train SliceNet. 

\subsection{SliceNet Dataset}

For the SliceNet dataset, we collected separate data for each of the following event classes for monitoring slicing:

\begin{enumerate}
    \item Hitting the cutting board: The knife robot hit the cutting board at random locations 60 times.
    \item Scraping the cutting board: The knife robot scraped the cutting board at random locations 10 times.
    \item Hitting an object: The knife robot hit each type of object between 10 and 15 times depending on the length of the object.
    \item Scraping an object: The knife scraped each object twice, once from left to right and once from right to left a distance between 5cm to 10cm depending on the length of the object.
    \item Slicing an object: The knife robot executed between 20 and 40 DMP slicing actions on each object depending on the thickness of the object, resulting in around 10 to 15 slices cut from each object. 
    \item In the air (Background): The knife robot executed DMP slicing actions 10 times at random locations in the air. 
\end{enumerate}

\begin{figure}[t]
    \centering
    \includegraphics[width=0.48\textwidth]{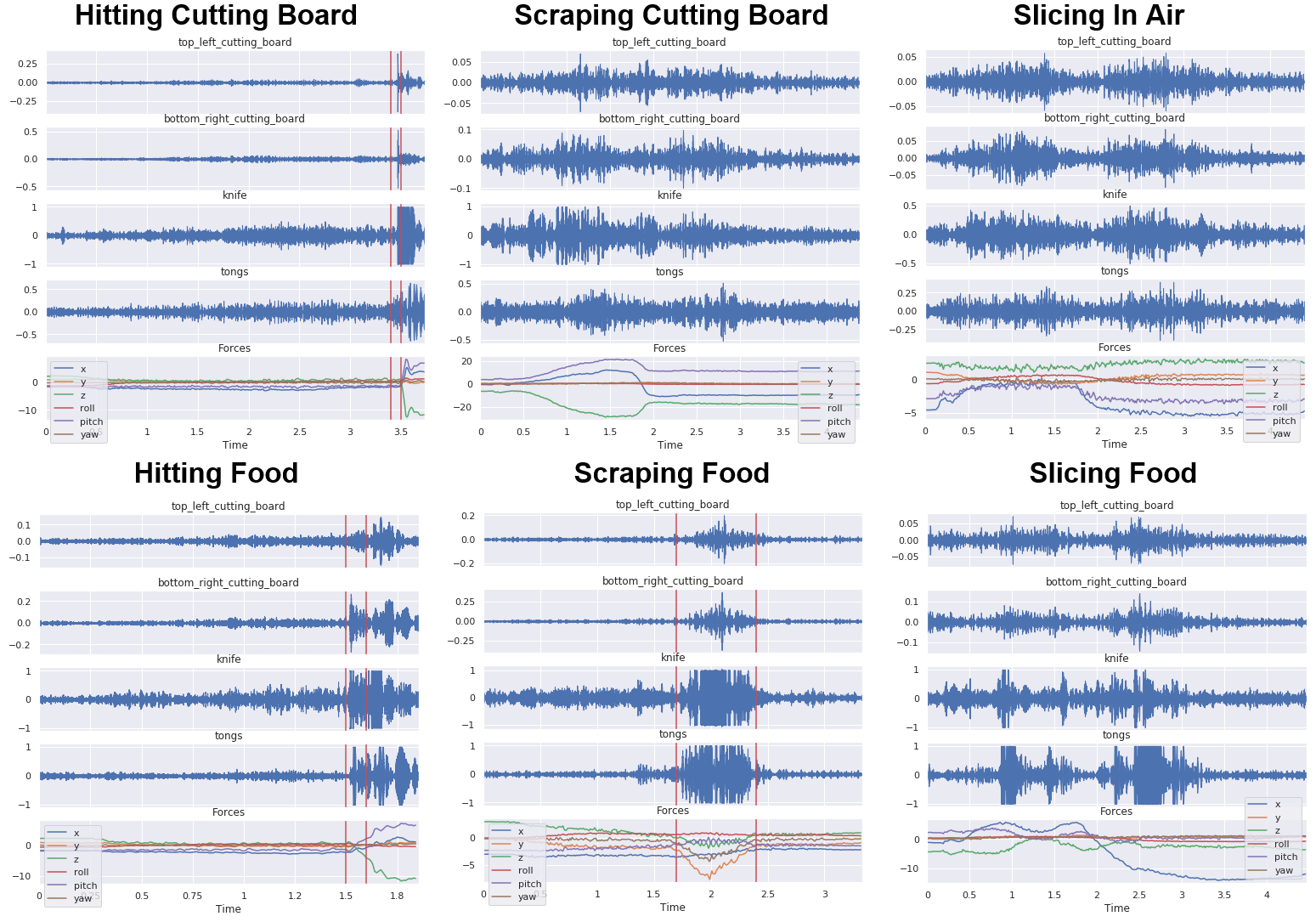}
    \caption{Sound signals from each of the 6 events that may occur when cutting.}
    \label{fig:sound_events}
    \vspace{-1em}
\end{figure}

Example vibration signals from each event class are shown in Fig.~\ref{fig:sound_events} above. To label the data, we segmented the vibration signals based on the skill the robot was executing. We then used an online Bayesian changepoint detection algorithm \cite{adams2007bayesian} and force gradient thresholding to segment out the vibrations of the actual event. For example, when the robot was moving down to hit the cutting board, we knew that the robot was in the air until a change was detected in both sound and forces. For skills where the robot remained within a specific contact state, we simply labeled all of the windows from that skill execution the same. For example, if the robot was still above the cutting board when executing a DMP slicing skill, we know that it was entirely within the slicing-an-object state.

Using these labelling methods, we constructed the SliceNet dataset. The red vertical lines in Fig. \ref{fig:sound_events} signify the changepoints that were detected by the online Bayesian changepoint detection algorithm.

\subsection{FoodNet Dataset}
\begin{figure}[t]
    \centering
    \includegraphics[width=0.48\textwidth]{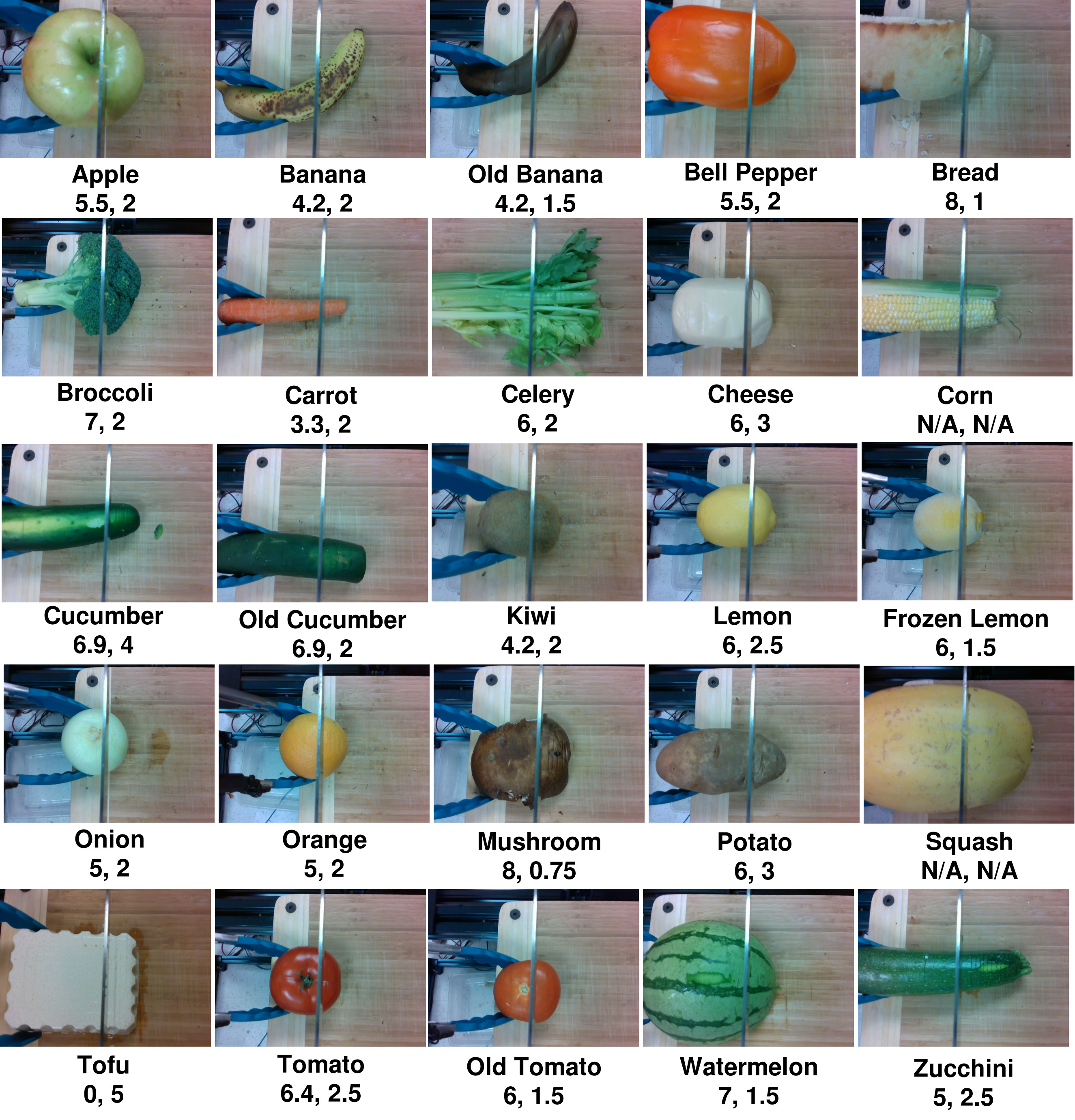}
    \caption{Images of all the Food Items and their DMP Parameters $\phi_1^x$ and $\phi_1^z$.}
    \label{fig:food_collage}
    \vspace{-4mm}
\end{figure}

For training FoodNet, we reused the hitting data from the SliceNet dataset, but we labeled skills based on the food type. In addition, we collected empirical data on the $\phi_1^x$ and $\phi_1^z$ parameters of the slicing DMP that could cut the objects fast and robustly. Although not optimal, these parameter values worked considerably better and faster than the initial constant DMP parameters. The full list of food items and their DMP parameters are illustrated in Fig. \ref{fig:food_collage}. Old food items refer to ones that were in the refrigerator for longer than a week.


\section{Experiments} \label{sec:experiments}
We now present results for both multimodal contact event monitoring using the SliceNet dataset and dynamic slice adaptation using the FoodNet dataset. In both of these settings, we show that our system is reliably able to classify the different events and the material properties of the object from the multimodal input data. Finally, to verify if both modalities are useful, we perform an extensive ablation study to compare the performance across different inputs.

\subsection{SliceNet}

\begin{figure}[t]
\centering
\begin{minipage}{.23\textwidth}
  \centering
  \includegraphics[width=\linewidth]{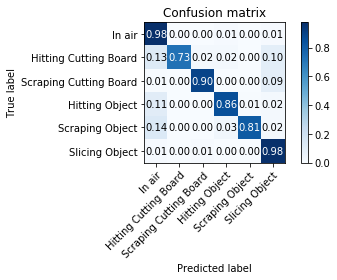}
    \caption{SliceNet Normalized Confusion Matrix}
    \label{fig:slicenet_confusion_matrix}
\end{minipage}
\hfill
\begin{minipage}{.23\textwidth}
  \centering
  \includegraphics[width=\linewidth]{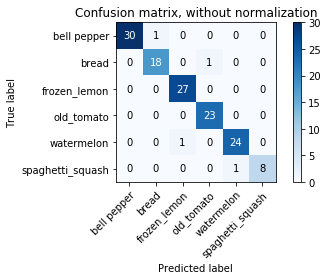}
  \caption{FoodNet Confusion Matrix}
  \label{fig:foodnet_confusion_matrix}
\end{minipage}
\end{figure}

\begin{figure}[t]
\centering
\begin{minipage}{.23\textwidth}
  \centering
  \includegraphics[width=\linewidth]{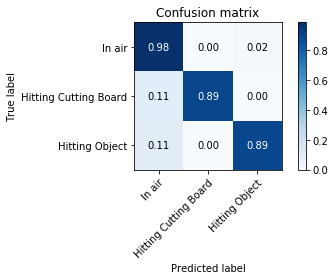}
  \caption{Hitting Confusion Matrix}
  \label{fig:hitting_confusion_matrix}
\end{minipage}
\hfill
\begin{minipage}{.23\textwidth}
  \centering
  \includegraphics[width=\linewidth]{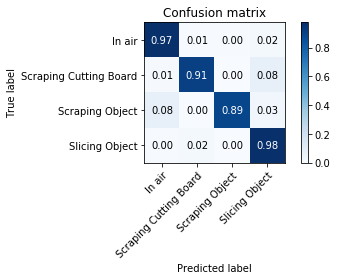}
  \caption{Slicing Confusion Matrix}
  \label{fig:slicing_confusion_matrix}
\end{minipage}
\vspace{-0.5em}
\end{figure}

To test the accuracy of our SliceNet model, we divide the SliceNet dataset into a train-test split, wherein we use 20\% of the data for the test split. 
Using our SliceNet architecture, we get a weighted F1-score of 0.959.
Fig.~\ref{fig:slicenet_confusion_matrix} shows the normalized confusion matrix for predicting the different events during the slicing DMP action. Our network is correctly able to classify most of the events. However, there does exist some confusion between scraping an object and being in air. We believe this is because there is often only a faint scraping sound from the knife, which is not captured by the vibration features or forces. Furthermore, there also exists some confusion between hitting the object and being in the air, which we believe is due to soft hits on objects such as oranges or lemons, which dampen the sound and increase the forces slowly. Finally, we have some confusions between hitting the cutting board, scraping the cutting board, and slicing objects. There is little confusion between other events, which shows that our network can accurately classify those events. 

Because of the confusions in SliceNet, we also trained separate networks, with the same architecture, for the hitting events and slicing actions respectively. Since these events are mutually exclusive, training separate networks should provide additional context and remove some confusion from the network. Fig.~\ref{fig:hitting_confusion_matrix} shows the results for the hitting events while Fig.~\ref{fig:slicing_confusion_matrix} shows the confusion matrix for the slicing events. For both of these classes we get an F1-score of 0.9685 and 0.9661 respectively. The improved performance of separate models shows that there might be some similarities in the vibration features and forces between hitting events and slicing actions which can reduce the performance of a single network trained to classify all events.


\subsection{FoodNet}

\begin{figure}[t]
    \centering
    \includegraphics[width=0.47\textwidth]{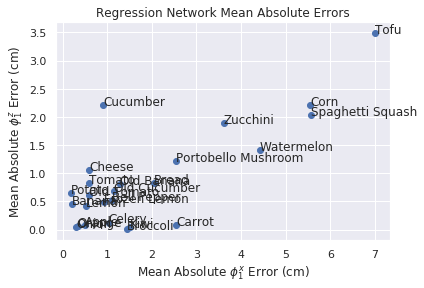}
    \caption{FoodNet Regression Network Leave-one-out Cross-Validation Mean Absolute Error in $\phi_1^x$ and $\phi_1^z$ }
    \label{fig:regression_network_plot}
    \vspace{-4mm}
\end{figure}


Our FoodNet architecture was able to achieve a F1-score of 0.9915 on a held out test set of 20\% of the total data. Its confusion matrix with only the classes that were confused is shown in Fig. \ref{fig:foodnet_confusion_matrix}. Frozen lemon, watermelon, and spaghetti squash were confused most likely due to their relative hardness. Meanwhile, bread was confused with both bell pepper and old tomatoes, perhaps due to the similar bouncy nature of their skins. FoodNet was able to detect the material differences between every other class without errors.

To test generalization to unknown materials, we trained an additional regression network with the same neural network architecture except with ReLu activations that outputs the parameters $\phi_1^x$ and $\phi_1^z$ directly. Instead of solely using the original FoodNet dataset that only shared hitting data with the SliceNet dataset, we augmented it with the DMP slicing actions in the SliceNet dataset, which allowed the network to predict the parameters continuously. The network had a mean absolute error of 0.019cm in $\phi_1^x$ and 0.013cm in $\phi_1^z$ on a test set of 20\% of the total data. 

We then perform leave-one-out cross-validation for each class of object. We calculated the Mean Absolute Error for both $\phi_1^x$ and $\phi_1^z$ of each class and plotted it above in Fig. \ref{fig:regression_network_plot}. As shown in the figure, certain classes are outliers. When these classes are left out, the network has large prediction errors such as Tofu, which was the only object that could be cut by just moving straight down. Corn and spaghetti squash were both unable to be cut, so they had large errors as we labeled both of their $\phi_1^x$ and $\phi_1^z$ parameters as 0. Since watermelon was hard like corn and spaghetti squash, it is likely that watermelon's predicted values were influenced by their labeled parameters, which resulted in large errors. Interestingly, cucumber had a large mean absolute $\phi_1^z$ error because the network predicted the $\phi_1^z$ value of old cucumber for the regular cucumber which is 2cm off in the $Z$-axis. The rest of the objects were clustered closer to the origin and the slight errors were negligible for real-world usage as the DMP action parameters in Fig.~\ref{fig:food_collage} are not optimal.

\subsection{Ablation Studies}
\label{sec:ablation_study}

\begin{table}[t]
\centering
\begin{tabular}{ l c c c } 
 \toprule
 Inputs & SliceNet & FoodNet \\
 \midrule
 Combined Sound and Force Features & 0.957 & 0.984 \\
 Forces & 0.927 & 0.179 \\
 Sound & 0.936 & 0.983 \\
 Under Cutting Board Mic 1 & 0.820 & 0.713 \\
 Under Cutting Board Mic 2 & 0.836 & 0.734 \\
 Knife Arm Mic 3 & 0.907 & 0.615 \\
 Tong Arm Mic 4 & 0.897 & 0.936 \\
 MFCC & 0.935 & 0.974 \\
 Chroma & 0.749 & 0.457 \\
 Mel & 0.935 & 0.960 \\
 Spectral & 0.719 & 0.305 \\
 Tonal & 0.536 & 0.056 \\
 Mfcc and Forces & \textbf{0.959} & \textbf{0.991} \\
 \bottomrule
\end{tabular}
\caption{F1-scores with Various Features}
\label{tab:feature_accuracies}
\vspace{-2em}
\end{table}

To verify how the network performance varies across different inputs, we evaluated each type of feature as well as each microphone individually. Table~\ref{tab:feature_accuracies} shows the F1-score for all of the different input features being evaluated. Using force feedback alone leads to a good performance for classifying events, but it performs poorly when classifying the different food items. 

By contrast, using only vibration data still shows very good performance. This clearly shows that vibrational data is more discriminatory for both contact event detection and material classification.
Among the different microphones, we observe that using only one of the microphones does reduce the classification accuracy. Thus, a distribution of microphones scattered through the scene improves performance by a large margin.
The results show that the contact microphone on the arm with the tongs is the most distinguishable for FoodNet. We believe that this occurs because 
the robot is always vibrating due to its motors, but when the robot grasps a food item, the sound is noticeably dampened, which allows the microphone to capture the discriminatory sounds before beginning the slicing skill.

Among the different types of acoustic features, we observe that using the Mel Frequency Cepstral Coefficient (MFCC) features are sufficient to achieve good performance for both SliceNet and FoodNet. We believe this is because the MFCC features capture the most salient parts of the vibrational feedback while discarding the background noise and thus are the most useful for our classification tasks.

\subsection{Robot Experiments}

We conducted a baseline comparison of using FoodNet Parameters vs a slicing DMP without adaptation on 7 items that were in our training set and 3 novel items that were not in our training set. We present the average time it takes to create a slice over 5 trials in Table \ref{tab:slicing_time_table}. In some cases, there is a neglible time difference when using FoodNet parameters, while in other cases, the average time decreases by a significant amount because the food items are easier to cut. Thus, a single slicing motion can complete a slice instead of multiple conservative slicing motions.

In addition, when we experimented with new novel items, FoodNet was able to output DMP slicing parameters similar to those of materials with similar textures in the training data. For example, a unique Peach was classified to have a similar texture to Zucchinis, Pears were similar to Apples, and finally Jalape\~{n}o Peppers were classified to be similar to Bell Peppers. In our accompanying video located here: \url{https://youtu.be/WgoAuyR31dY}, we show the experimental results. 
When the Knife Arm reached the center of a Peach's core in our tests, it did trigger the scraping cutting board classification due to the high forces and relatively similar sounds. In order to cut peaches and avocados, we will need to develop novel regrasping techniques to rotate the food item in order to cut it in half and remove the cores as our robot is unable to cut straight through them.

\begin{table}[t]
\centering
\begin{tabular}{ l c c c } 
 \toprule
 Food Item & Without Adaptation & FoodNet & Change \% \\
 \midrule
 Banana & 3.92 & 4.02 & \textbf{+2.56} \\
 Broccoli & 11.47 & 9.42 & \textbf{-17.88} \\
 Carrot & 3.98 & 2.03 & \textbf{-48.95} \\
 Cucumber & 5.95 & 2.01 & \textbf{-66.23} \\
 Cucumber (old) & 3.99 & 3.97 & \textbf{-0.64} \\
 Jalape\~{n}o Pepper & 2.83 & 2.07 & \textbf{-26.93} \\
 Kiwi & 5.61 & 3.94 & \textbf{-29.80} \\
 Peach & 7.90 & 5.56 & \textbf{-29.54} \\
 Pear & 6.37 & 5.16 & \textbf{-18.91} \\
 Tomato & 7.89 & 5.57 & \textbf{-29.32} \\
 \bottomrule
\end{tabular}
\caption{Average Time in Seconds to Slice each Food Item over 5 Trials using FoodNet or Without Adaptation.}
\label{tab:slicing_time_table}
\vspace{-1em}
\end{table}



\section{CONCLUSIONS AND FUTURE WORK}


We have presented a robust slicing approach that successfully utilizes both vibration and force feedback to adapt the cutting motions. We proposed one neural network (SliceNet) to continually monitor the robot's contact state and another (FoodNet) to predict the type of food item. The predicted food type is then used to automatically set the parameters of an adaptive slicing DMP. The proposed approach was successfully implemented on a real robot. Our experiments show that the framework allows the robot to adapt to a wide variety of food items, cutting items faster and more reliably.




In the future, we plan to explore different styles of cutting, such as chopping small items or carving out the solid cores of peaches and avocados. We will also use reinforcement learning to acquire the DMP parameters for new types of food items autonomously.


\bibliography{main} 
\bibliographystyle{ieeetr}

\end{document}